\documentclass{article}

    \usepackage[preprint, nonatbib]{neurips_2020}

\usepackage[utf8]{inputenc} 
\usepackage[T1]{fontenc}    
\usepackage{hyperref}       
\usepackage{url}            
\usepackage{booktabs}       
\usepackage{amsfonts}       
\usepackage{nicefrac}       
\usepackage{microtype}      
\usepackage{amsmath}
\usepackage[dvipsnames]{xcolor}
\usepackage{tabularx}
\usepackage{algorithm} 
\usepackage{algpseudocode}
\usepackage{graphicx}
\usepackage{mathtools}
\usepackage{multicol}
\usepackage{subcaption}
\usepackage{wrapfig}

\usepackage{amsthm}

\newcommand{\R}{\mathbb{R}}

\DeclareMathOperator*{\argmin}{arg\,min}

\newcommand{\ours}{Geodesic-HOF}

\title{Geodesic-HOF: 3D Reconstruction Without Cutting Corners}

\author{%
  Ziyun Wang, Eric A. Mitchell, Volkan Isler, Daniel D. Lee \\
  Samsung AI Center\\
  New York, NY 10011 \\
  \texttt{saicny@samsung.com}
}

\begin{document}

\maketitle

\begin{abstract}

Single-view 3D object reconstruction is a challenging fundamental problem in computer vision, largely due to the morphological diversity of objects in the natural world. In particular, high curvature regions are not always captured effectively by methods trained using only set-based loss functions, resulting in reconstructions short-circuiting the surface or cutting corners. In particular, high curvature regions are not always captured effectively by methods trained using only set-based loss functions, resulting in reconstructions short-circuiting the surface or cutting corners. To address this issue, we propose learning an image-conditioned mapping function from a canonical sampling domain to a high dimensional space where the Euclidean distance is equal to the geodesic distance on the object. The first three dimensions of a mapped sample correspond to its 3D coordinates. The additional lifted components contain information about the underlying geodesic structure. Our results show that taking advantage of these learned lifted coordinates yields better performance for estimating surface normals and generating surfaces than using point cloud reconstructions alone. Further, we find that this learned geodesic embedding space provides useful information for applications such as unsupervised object decomposition.



\end{abstract}
\section{Introduction}

Reconstructing the 3D model of an object from a single image is a central problem in computer vision, with many applications. For example, in computer graphics, surface models such as triangle meshes are used for computing object appearance. In robotics, surface normal vectors are used for grasp planning, and in computer aided design and manufacturing, complete object models are needed for production. Motivated by such use cases, recent deep learning-based approaches to 3D reconstruction have shown exciting progress by using powerful function approximators to represent a mapping from the space of images to the space of 3D geometries. It has been shown that deep learning architectures, once trained on large datasets such as ShapeNet~\cite{chang2015shapenet}, are capable of outputting accurate object models in various representations including discrete voxelizations, unordered point sets, or implicit functions.

\begin{figure}
    \centering
    \includegraphics[width=\textwidth]{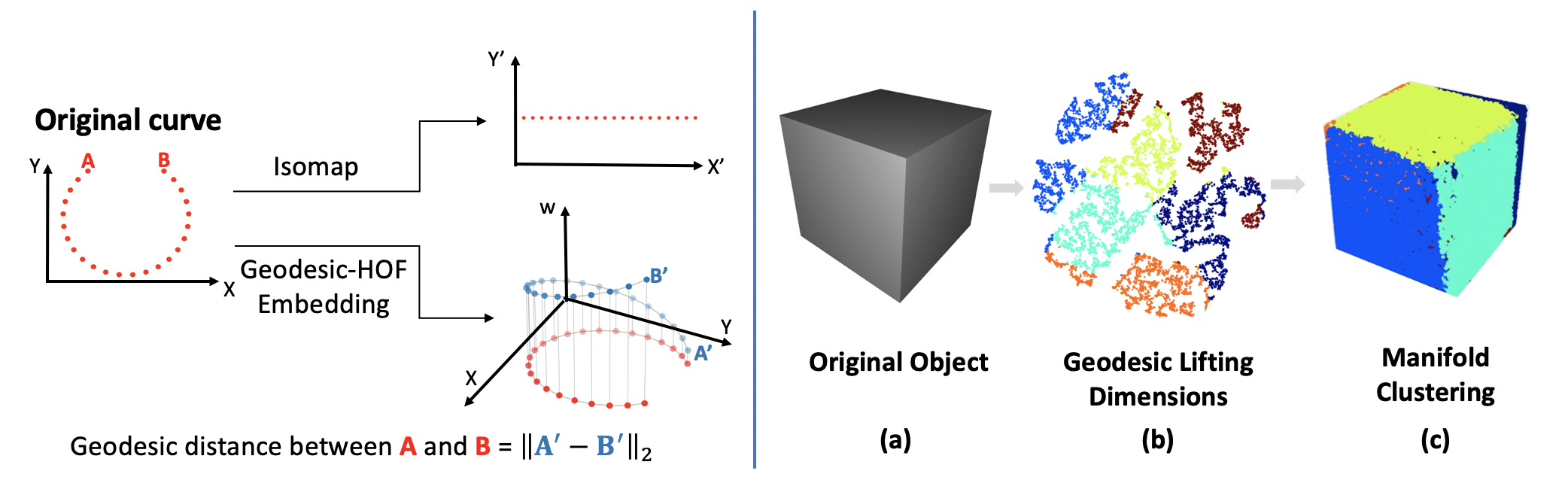}
    \caption{\textbf{Left:} Embedding a cut circle with Isomap~\cite{tenenbaum2000global} vs. \ours{}. In contrast to embedding methods such as Isomap~\cite{tenenbaum2000global}, \ours{} preserves the original Euclidean coordinates and lifts the points to the higher-dimensional space by adding extra "geodesic lifting coordinates". \textbf{Right:}~Face Decomposition using the geodesic lifting coordinates. Figure~(a) shows the object of interest. Figure~(b) shows a low-dimensional projection of the geodesic lifting dimensions using t-SNE~\cite{maaten2008visualizing}. Figure~(c) shows the same clustering results by coloring the output 3D points. For clarity of correspondence, we retrospectively color the points in (b) to match the clustering labels in (c).  Best viewed in color.}
    \label{fig:overview}
    \vspace{-10pt}
\end{figure}

These representations can generally produce aesthetically pleasing reconstructions, but extracting topological information from them, such as computing neighborhoods of a point on the surface, can be difficult. In particular, naively computing a neighborhood of a point using a Euclidean ball or Euclidean nearest neighbors can give incorrect results if the object has regions of high curvature. Such mistakes can in turn degrade performance on downstream tasks such as unsupervised part detection or surface normal estimation. 
We thus propose a new method for single-view 3D reconstruction based on the idea of explicitly learning surface geodesics. The key insight of our approach is to embed points sampled from the surface of a 3-dimensional object into a higher-dimensional space such that geodesic distances on the surface of the object can be computed as Euclidean distance in this `lifted' space. This embedding space is constrained such that the 3D surface of the object lies in its first 3 dimensions (Figure~\ref{fig:overview}-left shows a 2D example of this constraint). The remaining embedding dimensions can thus be interpreted as a quantification of curvature. To test the efficacy of this approach, we present a neural network architecture and training regime that is able to effectively learn this representation, and our experiments demonstrate that using the learned surface geodesics yields meaningful improvements in surface normal estimation. Further, we find that this embedding enables unsupervised decomposition of an object's surface into non-overlapping 2D sub-manifolds (i.e., charts), which can be useful for texture mapping and surface triangulation (see Figure~\ref{fig:overview}-right).

\paragraph{Our contributions.} We present \ours{} for surface generation and provide several experimental evaluations to assess its  performance. In Section~\ref{sec:svr}, we show that \ours{} exhibits reconstruction quality on par with state-of-the-art surface reconstruction methods. Further, we find that using the learned surface geodesics meaningfully improves surface normal estimation. Finally, in Section~\ref{sec:manifold}, we show how the learned representation can be used for decomposing the object surface into non-overlapping charts which can be used for generating triangulations or explicit function representations of surfaces. We hope that our results will provide an effective method for reconstructing surfaces and unfold new research directions for incorporating higher-order surface properties for shape reconstruction.

\section{Related Work}

Several recently-developed object representations have shown promise in the 3D reconstruction literature. With the success of Convolutional Neural Network (CNN) in processing image data, it is natural to extend it to 3D. Therefore, many methods have been developed to directly output a regular voxel grid in 3D \cite{choy2016r2n2,tatarchenko2017octree,riegler2017octnet,wu2016learning}. However, the naive voxel representation requires the output to be the same resolution during training and during inference and the computational and memory usage of such methods grows cubically with the resolution. More complex octree-style approaches have been proposed to address these issues, but scaling to high resolutions remains a challenge~\cite{riegler2017octnet}.

In light of these resource demands, unordered point sets have become a popular alternative to voxelization for representing 3D shapes~\cite{fan2017point,yang2018foldingnet,lin2018learning,mitchell2019higher}, as the inherent sparsity of point sets scale more gracefully to high-resolution reconstructions. However, unordered point sets still lack intrinsic information about the topology of the underlying surface; thus, some work has investigated the use of implicit functional surface representations. Implicit functions such as occupancy~\cite{mescheder2019occupancy} or signed distance~\cite{park2019deepsdf}, which are continuous by definition, have shown promise as object representations, and can effectively store detailed information about an object's geometry. One of the main drawbacks of such methods is the need for a post-processing step such as running marching cubes to generate the object, making extracting the underlying object extremely time-consuming~\cite{park2019deepsdf}. Furthermore, generating training data for these methods is non-trivial since they require dense samples near the surface of the object. Finally, as noted in~\cite{mescheder2019occupancy}, in terms of Chamfer score, implicit function methods are not as accurate as \emph{direct methods} such as AtlasNet~\cite{atlasnet} and Pixel2Mesh~\cite{wang2018pixel2mesh} which are trained directly using  Chamfer-based loss functions.

A substantial body of work exists at the intersection of differential geometry, computer vision, and deep learning that studies the usefulness of geodesics in many 3D settings, such as surface matching~\cite{wang2000shape,Wang2003}, shape classification~\cite{Luciano2018}, and manifold learning \cite{pai2019dimal,atlasnet,wang2018pixel2mesh}.
In particular, the well-known Isomap~\cite{tenenbaum2000global} algorithm uses shortest paths on k-nearest neighborhood graphs and applies Multi-Dimensional Scaling~\cite{cox2008multidimensional} to find the dimensionality and embedding of the data. As illustrated in Figure~\ref{fig:overview}, direct embedding of geodesic distances does not necessarily yield the object surface. In \ours{}, the network is designed to explicitly output a sampling of the surface manifold and learn geodesic distances.

\section{Learning Geodesics for 3D Reconstruction}
\label{sec:ours}

\begin{figure}
    \centering
    \includegraphics[width=\textwidth]{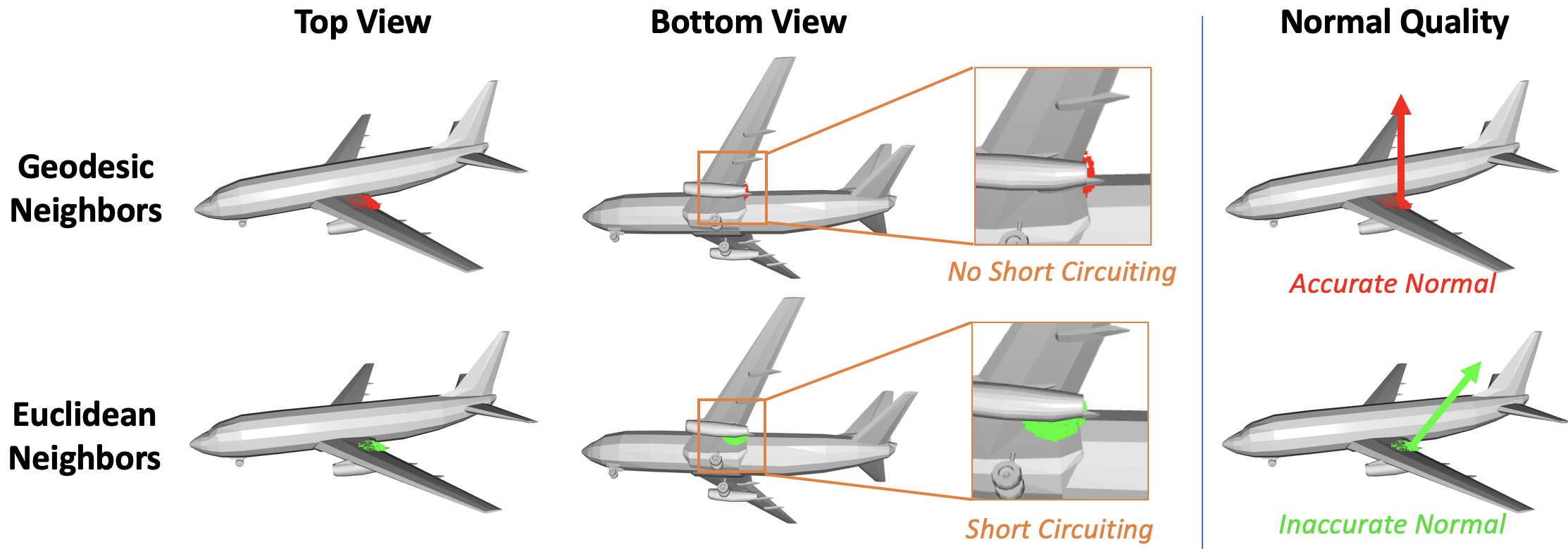}
    \caption{Normal Estimation with Geodesic Neighborhoods vs. Euclidean Neighborhoods. \textbf{Top:}~Estimating the normal of a point on the wing by using geodesic neighborhoods. \textbf{Bottom:} Normal estimation of the same point using the Euclidean neighborhoods. We observe spurious neighbor points from the bottom of the wing are included in the Euclidean case, which causes the normal estimation to be inaccurate. Best viewed in color.}
    \label{fig:norm_est}
    \vspace{-15pt}
\end{figure}

From a finite set of points, connecting points based on Euclidean distance proximity alone is insufficient to produce an accurate depiction of the surface topology. If distant points on the manifold are erroneously considered to be close because they are close in the Euclidean space used for computing neighborhoods, the so-called ``short-circuiting" problem arises~\cite{balasubramanian2002isomap}.
Short-circuiting can be observed in Figure~\ref{fig:norm_est} where the points on opposite sides of a single wing are erroneously connected because they are nearby in terms of their Euclidean distance, although they are quite far on the surface. We propose using surface geodesics as a natural tool to solve this problem.

A \emph{geodesic} between two points on a surface is a shortest path between these points which lies entirely on the surface\footnote{In some contexts, geodesics are defined as locally shortest paths. For example, two points on the sphere making an angle less than $\pi$ with the center has two geodesics even though one is shorter than the other. In this paper, we use the term to refer to a globally shortest path. Note that, there might still be multiple geodesic as in the case of two diametrically opposite points on the sphere.}. Geodesics carry a significant amount of information about surface properties. In particular, the difference between the geodesic and Euclidean distances between two points is directly related to curvature. Intuitively, connecting points based on geodesic distance, rather than Euclidean distance yields a more faithful reconstruction of the true surface. Given this setup, we can formalize the surface reconstruction problem that is the focus of this work.

\textbf{Problem statement:} Given a single image $I$ of an object $O$, our goal is to be able to (i)~generate an arbitrary number of samples from the surface of $O$, and (ii)~compute the geodesic distance between any two generated points on the surface. We use Chamfer distance (Eqn.~\ref{eqn:chamfer}) to quantify the similarity between a set of samples from the model and the ground truth $O$.

In the next section,  we present our approach to solving this problem through a deep neural network.

\subsection{\ours{}: Method Overview}

\ours{} is a neural network architecture and set of training objectives that aim to address the problem presented above. In order to describe the technical details of \ours{}, we first establish our notation and terminology. During training, we are given an image $I$ of an object $O$ as well as $X^* = \{x^*_1, \ldots, x^*_n\}$, a set of points sampled from $O$. We denote the ground truth surface geodesic function $g(\cdot, \cdot)$. In practice, the function $g$ can be computed using either a very dense sampling of the ground truth object $O$ or a surface triangulation. Using a convolutional neural network, \ours{} maps an input image $I$ to a continuous mapping function $f_I: M \subset \R^D \rightarrow \R^{3+K}$. $f_I$ maps samples from a canonical sampling domain $M$ 
to an embedding space $Z = \{z_i = f_I(m_i)\in \R^{3+K}\}$. In this work, we take $M$ to be the unit solid sphere.




\newcommand{\hz}[0]{\hat{z}}
\newcommand{\hx}[0]{\hat{x}}

 
For every $z \in Z$, we denote the vector obtained by taking the first three components of $z$ as $x$ and refer to it as \textit{point coordinates} of $z$. We call the remaining $K$ dimensions as the \textit{geodesic lifting coordinates} of $z$. We define $\hat{O} = \{\hx_i = f_I(m_i)\}$ be the set of predicted point coordinates from a set of samples $\{m_i\}$. Finally, for any two $m_i, m_j \in M$, with $\hz_i =f_I(m_i) $ and $\hz_j = f_I(m_j)$, 
the predicted geodesic distance is given by $\hat{g}(\hz_i, \hz_j) = || \hz_i - \hz_j||_2$.

The mapping $f_I$ is trained such that $\hat{O}$ accurately approximates the ground truth object $O$ in terms of Chamfer distance and $\hat{g}(z_i, z_j)$ accurately approximates the ground truth geodesic distance $g(x^*_i, x^*_j)$, where $x^*_i, x^*_j \in X^*$ are the two ground truth samples closest to $\hx_i$ and $\hx_j$ (the point coordinates of $z_i$ and $z_j$. The first condition requires that the point coordinates represent an accurate sampling of the object surface; the second condition requires the embedding space to accurately capture geodesic distances between two point coordinates. We can show that the lifting coordinates encode curvature information: the quantity $\hat{g}(\hz_i,\hz_j)^2 -||\hx_i - \hx_j||^2$  is equal to the squared norm of the geodesic lifting coordinates. This quantity approaches the geodesic curvature when the samples are close to each other on the manifold.
 


\subsection{Network Architecture and Training Procedures}

The network design of \ours{} follows the \textbf{Higher Order Function (HOF)} method, which has three main components: an image encoder, a Higher-Order Function network and a point mapping network. An image encoder extracts the semantic and geometric information from an image $I$ of an object. From this representation, the Higher-Order Function network predicts a set of parameters of a mapping function $f_I: \R^D \rightarrow \R^{3+K}$. To use the mapping function $f_I$, we start by sampling the unit sphere to generate a set of points $M = \{m_i\}$. Then we use the learned network to map these sampled points to a set of embeddings $Z = \{z_i = f_I(m_i)\in \R^{3+K}\}$. The advantages of Higher-Order Function Networks over the Latent Vector Concatenation (LVC) paradigm are discussed in detail in \cite{mitchell2019higher}. Please refer to the Supplementary Material Section for the architecture details. In \ours{}, we optimize the loss function $L$, the weighted sum of the standard Chamfer loss $L_C$ and the geodesic loss $L_G$, with weight $\lambda_C$ and $\lambda_G$, respectively
\begin{align}
    L := \lambda_\text{C} L_\text{C} + \lambda_\text{G} L_\text{G}
\end{align}
These losses are defined in the next section. Since the precision of the point coordinates is important for finding the correct geodesic distance, we weigh the Chamfer loss more than the Geodesic loss. Practically, we choose $\lambda_\text{G}$ and $\lambda_\text{C}$ to be 0.1 and 1.0 respectively. We use the Adam Optimizer~\cite{kingma2015adam} with learning rate $1$e-5.

\subsection{Loss Functions}
The weights of the mapping function $f_I$ for a given image $I$ are learned so as to minimize the weighted sum of two losses: \textit{Chamfer loss} and \textit{Geodesic loss}. Recall $M$ is a set of points randomly sampled from the pre-mapping space. The predicted set of embeddings from $M$ is $Z = \{z_i = f_I(m_i) \mid m_i \in M\}$.
 For simplicity of notation, we separate the 3D point coordinates $X = \{x_i \in \R^3\}$ and the geodesic lifting coordinates $W = \{w_i \in \R^K\}$ such that $z_i \coloneqq [x_i; w_i]$.
\paragraph{Chamfer loss} is defined as the set distance between the point coordinates $X=\{x_i\}$ and the ground truth point set $Y$.
\begin{align}
L_{\text{C}}(X, Y) = \frac{1}{|X|} \sum_{x \in X} \min_{y \in Y} ||x - y||^2_2 + \frac{1}{|Y|} \sum_{y \in Y} \min_{x \in X} ||y -x||^2_2
\label{eqn:chamfer}
\end{align}
We optimize this loss so that our predicted point set accurately represents the surface of the ground truth object. 
\paragraph{Geodesic loss} ensures the geodesic distance is learned accurately between every pair of points. We denote the ground truth geodesic distance on the object surface as $g(x_i, x_j)$, where $(x_i, x_j) \in \hat{O}$ and $\hat{O}$ is our prediction of $O$, which is the union of several 2-manifolds representing the object surface in $\R^3$. Since the geodesic distance is only defined on the object surface, we project the point coordinates $\{x_i\}$ onto the object and compute the pair-wise geodesic distance. The details of this formulation are described in Equation~\ref{eqn:geo_actual}. We want to reconstruct the object while learning the embedding of each point so that the geodesic distance in the 3D object space $\R^3$ is the same as the Euclidean distance in the embedding space $\R^k$. For a set of embeddings $Z = \{z_i \coloneqq [x_i, w_i]\}$, the geodesic loss is defined as:
\begin{align}
L_{\text{G}}(Z) = \frac{1}{|Z|^2}\sum_{i=1}^{|Z|}\sum_{j=1}^{|Z|} (||z_i - z_j||_2 - g(x_i, x_j))^2
\end{align}
For computing the geodesic loss, we need the ground truth geodesic distance matrix on the object $O$. We build a Nearest Neighborhood graph $G = (V, A)$ on $X^*$, a set of samples from $O$. We define $D(v_i, v_j)$, the geodesic distance between $v_i$ and $v_j$, as the length of the shortest path between $v_i \in V$ and $v_j \in V$ computed by Dijkstra's algorithm on $G$.
For each point $x_i$ from our prediction, we find its k-nearest neighbors, denoted as $\Lambda(x_i) = \{v^p_i\}$ where $p$ is the index of each neighbor. For a pair of point coordinates $(x_i, x_j)$, assume the set of nearest neighbors in $V$ of $x_i$ and $x_j$ are $\{v^p_i\}$ and $\{v^q_j\}$, respectively. Here, we use $\gamma_{pq}(x_i, x_j)$ to denote the unnormalized confidence score that path between $x_i$ and $x_j$ goes through $v^p_i$ and $v^q_j$. $\sigma$ here is a generic Gaussian radial basis function. We define $\alpha_{ij}$, the confidence of an undirected path between $v_i \in V$ and $v_j \in V$, as:
 \begin{align}
    \alpha_{pq}(x_i, x_j) &= \frac{\gamma_{pq}(x_i, x_j)}{\sum_{p \in |\Lambda(x_i)|, q \in |\Lambda(x_j)|}\gamma_{pq}(x_i, x_j)} \\
    \gamma_{pq}(x_i, x_j) &= \sigma(x_i, v^p_i) \sigma(x_j, v^q_j) \\
                          &= \text{exp}(-(||x_i - v_i^p||_2^2 + ||x_j - v_j^q||_2^2))
 \end{align}
 Essentially, the confidence of a path between $x_i$ and $x_j$ going through the two vertices $v_i^p$ and $v_j^q$ is the normalized similarity between ($x_i, x_j$) from our prediction and their possible corresponding vertices ($v_i^p, v_j^q$) in the graph measured by a radial basis kernel. Because of the normalization step, the confidence over all possible paths can be seen a probability distribution over which vertices $(v_i^p, v_j^q)$ to choose on the ground truth object.
 Thus, we can define the \textbf{soft} geodesic loss function as:
\begin{align}
    g(x_i, x_j) = \sum_{v_i^p \in \Lambda(x_i), v_j^q \in \Lambda(x_j)} \alpha_{pq}(x_i, x_j) D(v_i^p, v_j^q)
    \label{eqn:geo_actual}
\end{align}

\vspace{-15pt}
\section{Experiments \label{sec:experiments}}
\begin{table}
    \centering
    \setlength{\tabcolsep}{3pt}
    \begin{tabular}{|c | c c c c c c|}
    \hline
    Category &  3D-R2N2~\cite{choy2016r2n2} & PSGN~\cite{fan2017point} & Pix2Mesh~\cite{wang2018pixel2mesh} & AtlasNet~\cite{atlasnet} & OccNet~\cite{mescheder2019occupancy} & Ours\\
    \hline
    Airplane    &  0.227 & 0.137 & 0.187 & 0.104 &   0.147 &            \textbf{0.099}\\
    Bench       &  0.194 & 0.181 & 0.201 & 0.138 &   0.155 &            \textbf{0.122}\\
    Cabinet     &  0.217 & 0.215 & 0.196 & 0.175 &   0.167 &            \textbf{0.134}\\
    Car         &  0.213 & 0.169 & 0.180 & 0.141 &   0.105 &            \textbf{0.100}\\
    Chair       &  0.270 & 0.247 & 0.265 & 0.209 &   0.180 &            \textbf{0.173}\\
    Display     &  0.314 & 0.284 & 0.239 & 0.198 &   0.278 &            \textbf{0.193}\\
    Lamp        &  0.778 & 0.314 & 0.308 & 0.305 &   0.479 &            \textbf{0.229}\\
    Speaker     &  0.318 & 0.316 & 0.285 & 0.245 &   0.300 &            \textbf{0.206}\\
    Rifle       &  0.183 & 0.134 & 0.164 & 0.115 &   0.141 &            \textbf{0.096}\\
    Sofa        &  0.229 & 0.224 & 0.212 & 0.177 &   0.194 &            \textbf{0.162}\\
    Table       &  0.239 & 0.222 & 0.218 & 0.190 &   0.189 &            \textbf{0.145}\\
    Telephone   &  0.195 & 0.161 & 0.149 & 0.128 &   0.140 &            \textbf{0.109}\\
    Vessel      &  0.238 & 0.188 & 0.212 & 0.151 &   0.218 &            \textbf{0.137}\\
    \hline
    mean &  0.278 & 0.215 &  0.216 & 0.175 & 0.215 & \textbf{0.141}\\
    \hline
    \end{tabular}
    \vspace{5pt}
    \caption{Chamfer Comparison: \ours{} achieves state of the art performance in Chamfer distance. We sample 100,000 points on the object of interest and the output of each method to compute the Chamfer distance.}
    \label{tab:chamfer}
    \vspace{-10pt}
\end{table}

\begin{table}
    \centering
    \setlength{\tabcolsep}{3pt}
    \begin{tabular}{|c|c c c c|c c|}
    \hline
    Category &  3D-R2N2~\cite{choy2016r2n2} & Pix2Mesh~\cite{wang2018pixel2mesh} & AtlasNet~\cite{atlasnet} & OccNet~\cite{mescheder2019occupancy} & Ours (Euc) & Ours (Geo)\\
    \hline
    Airplane    &  0.629 & 0.759 & 0.836              & 0.840               & 0.846             & \textbf{0.863}\\
    Bench       &  0.678 & 0.732 & 0.779              & \textbf{0.813}      & 0.778             & 0.795\\
    Cabinet     &  0.782 & 0.834 & 0.850              & \textbf{0.879}      & 0.871             & 0.870\\
    Car         &  0.714 & 0.756 & 0.836              & \textbf{0.852}      & 0.819             & 0.827         \\
    Chair       &  0.663 & 0.746 & 0.791              & \textbf{0.823}      & 0.802             & 0.810\\
    Display     &  0.720 & 0.830 & 0.858              & 0.854               & 0.834             & \textbf{0.862}\\
    Lamp        &  0.560 & 0.666 & 0.694              & 0.731               & 0.712             & \textbf{0.732}\\
    Speaker     &  0.711 & 0.782 & 0.825              & 0.832               & 0.828             & \textbf{0.838}\\
    Rifle       &  0.670 & 0.718 & 0.725              & 0.766               & \textbf{0.797}    & \textbf{0.797}\\
    Sofa        &  0.731 & 0.820 & 0.840              & \textbf{0.863}      & 0.853             & 0.855\\
    Table       &  0.732 & 0.784 & 0.832              & \textbf{0.858}      & 0.827             & 0.841\\
    Telephone   &  0.817 & 0.907 & 0.923              & \textbf{0.935}      & 0.922             & 0.933\\
    Vessel      &  0.629 & 0.699 & 0.756              & \textbf{0.794}      & 0.774             & 0.790\\
    \hline
    mean &  0.695 &  0.772 & 0.811 & \textbf{0.834} & 0.818 & 0.828\\
    \hline
    \end{tabular}
    \vspace{5pt}
    \caption{Normal Comparison: \ours{} achieves competitive performance with state of the art methods in normal consistency. We sample 100,000 points on the object of interest and the output of each method to compute the normal consistency. Note that OccNet~\cite{mescheder2019occupancy} is an implicit method, which gives it advantages in normal estimation due to marching cubes post-processing.}
    \label{tab:normal}
    \vspace{-15pt}
\end{table}
In this section, we evaluate the effectiveness of our method on a single-view reconstruction task. We show that \ours{} is able to reconstruct 3D objects accurately while learning the geodesic distance. We evaluate \ours{} on the ShapeNet~\cite{chang2015shapenet} dataset and show \ours{} performs competitively in terms of Chamfer distance and in normal consistency comparing against the current state of the art 3D reconstruction methods. Then we show a set of interesting applications of our learned embeddings in tasks such as manifold decomposition and mesh reconstruction. Our experiments show that we can learn important properties such as curvature from the learned embeddings to render a better representation of the object shape.

\subsection{Single View Object Reconstruction \label{sec:svr}}
In this section, we evaluate the quality of our learned representation by performing a single-view reconstruction task on the the ShapeNet~\cite{chang2015shapenet} dataset. For fair comparison, we use the data split provided in \cite{choy20163d}. We use the pre-processed ShapeNet renderings and ground truth objects from \cite{mescheder2019occupancy}. For evaluation, we use two main metrics: Chamfer distance and Normal consistency. The detail of the dataset can be found in Supplementary Material.

\paragraph{Evaluation Metrics} Chamfer distance is defined identically as Equation~\ref{eqn:chamfer}. We sample 100,000 points from both our output representation and the ground truth point cloud. Note that the original objects are normalized so that the bounding box of each object is a unit cube and we follow the evaluation procedure of \cite{mescheder2019occupancy} to use $1/10$ of the bounding box size as the unit. Normal consistency between two normalized unit vectors $n_i$ and $n_j$ is defined as the dot product between the two vectors. For evaluating the surface normals, we first sample oriented points from the object surface, denoted as $X_{pred} = \{(\Vec{x_i}, \Vec{n_i})\})$ and the set of ground truth points and the corresponding normals  $X_{gt} = \{(\Vec{y_j}, \Vec{m_j})\})$. The surface normal consistency between two set of oriented points sample from the object, denoted as $\Gamma$, is defined as:
\begin{align}
\Gamma(X_{gt}, X_{pred}) &= \frac{1}{|X_{gt}|}\sum_{i \in |X_{gt}|}{|{\Vec{n_i}\cdot{\Vec{m}_{\theta(x, X_{pred})}}}|} \\
\theta(x, X_{gt}\coloneqq\{(\Vec{y_j}, \Vec{m_j})\})) &= \argmin_{j \in |X_{gt}|} ||x - y_j||_2^2 
\end{align}
In Table \ref{tab:chamfer} we show our Chamfer distance comparison with 3D-R2N2~\cite{choy2016r2n2}, PSGN (Point Set Generating Networks)~\cite{fan2017point}, Atlasnet~\cite{atlasnet} and Pixel2Mesh~\cite{wang2018pixel2mesh}. In Table~\ref{tab:normal}, we show the normal comparison results with the same set of methods. In Table~\ref{tab:chamfer}, it can be seen that \ours{} can accurately represent the object as indicated by the best overall chamfer performance. By learning a continuous mapping function between the mapping domain and the embedding domain, we allow high-resolution sampling of the embedding space to represent the object. In addition, by using a Higher-Order Function Network as the decoder, we avoid the dimensionality constraints of the latent vector code to provide a more direct way for image information to flow into the mapping function. In comparison, the methods that have pre-defined connectivity, such as Pix2Mesh~\cite{wang2018pixel2mesh} and AtlasNet~\cite{atlasnet}, suffer from the topological constraints of the sampling domain.

Table 2 shows that our normal estimation results compared with other methods. The last two columns show the normal consistency obtained with two methods. Both methods use a nearest neighbor search to find the local planar neighborhood and estimate the normals based on a principle axes analysis of the neighborhood. Ours (Euc) uses Euclidean distance to find the neighborhood whereas Ours (Geo) uses the learned geodesic embedding to find the neighborhood. The comparison between the last two columns shows that the geodesic embeddings provide a better guide for finding the low curvature neighborhood. For example, on the edge of the table, if we estimate the normal of the points on the tabletop near the edge, we should avoid the inclusion of the points on the side of the table. From our experiments, the geodesic neighborhoods yield overall better normal estimation results. Note that implicit methods such as Occupancy Network have a natural advantage in normal estimation due to the filtering effect of the marching cubes post-processing, which is advantageous for datasets with few high-frequency surface features. Despite this difference, our normal consistency performs competitively against state of the art methods. This result highlights the effectiveness of the learned geodesics in understanding the local structure, such as curvature, of the object.
\vspace{-5pt}
\section{Applications}
\begin{figure}
    \begin{subfigure}{.45\textwidth}
    \centering
    \includegraphics[width=\textwidth]{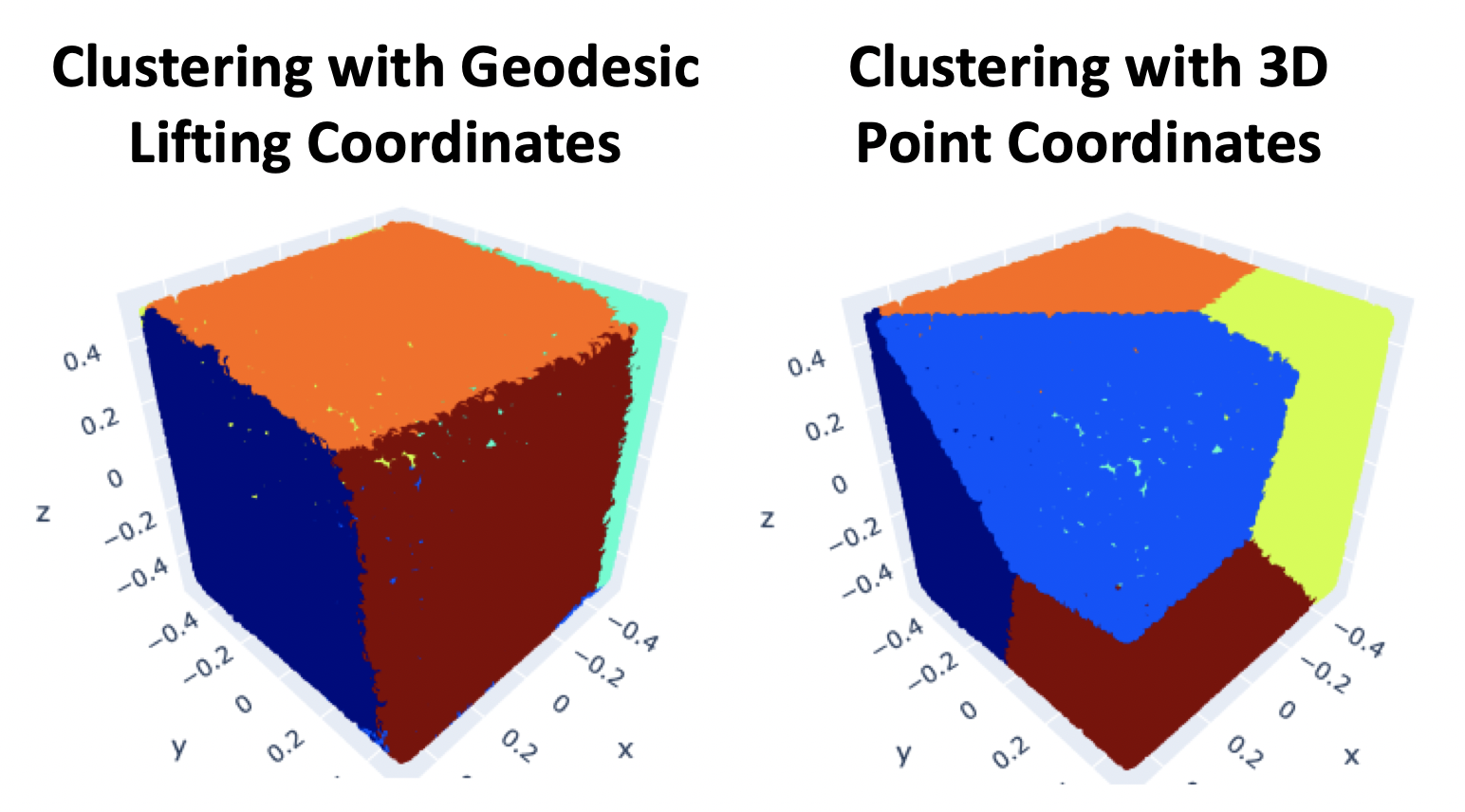}
    \caption{Manifold Decomposition of a canonical cube shape.}
    \end{subfigure} \hfill
    \begin{subfigure}{.45\textwidth}
    \centering
    \includegraphics[width=\textwidth]{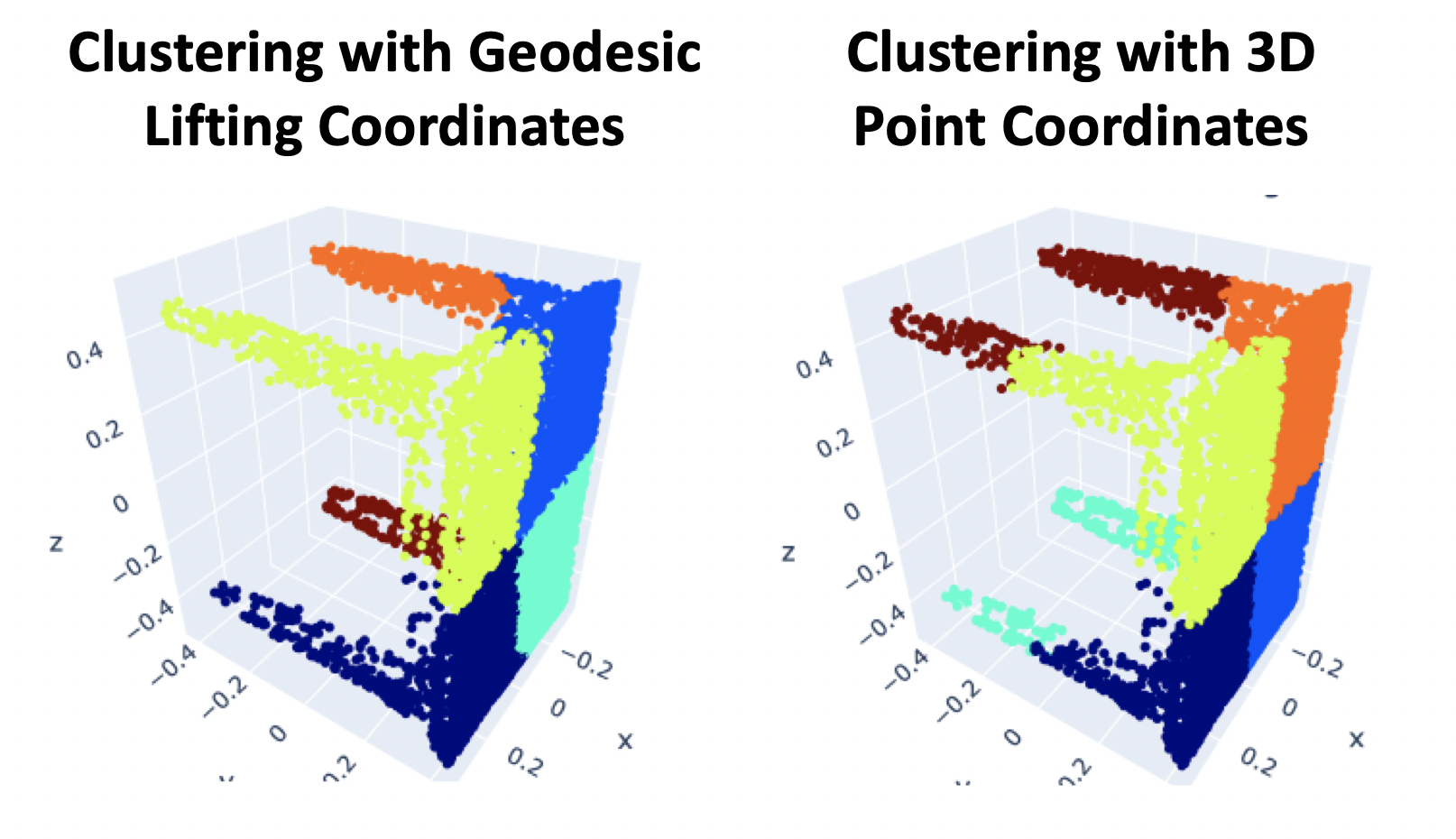}
    \caption{Manifold Decomposition of a table from the ShapeNet~\cite{chang2015shapenet} test set.}
    \end{subfigure}
    \caption{Comparing surface manifold decompositions by clustering raw 3D point coordinates $x_i$ and geodesic lifting coordinates $w_i$ of the predicted embedding $z_i=[x_i,w_i]$. K-Means is used for clustering with K=6. Geodesic lifting coordinates give more meaningful features that partition the surface into intuitive components. Best viewed in color.}
    \label{fig:clustering}
    \vspace{-15pt}
\end{figure}
In this section, we show two example applications of our learned geodesic embeddings. In Section~\ref{sec:manifold}, we use the geodesic lifting dimensions of the embeddings to decompose shapes into simpler, non-overlapping charts. In Section~\ref{sec:mesh}, we show how each chart from decomposition can be represented as an explicit function $y = f(u,v)$ which can then be used for mesh reconstruction.
\subsection{Chart Decomposition \label{sec:manifold}}
The object surface can be represented as a differentiable 2-manifold, also known as an atlas, using a collection of charts. Clustering \ours{} embeddings according to the lifting coordinates provides a natural way to split the object into a small number of low curvature charts  as illustrated in Figure~\ref{fig:clustering}, where we contrast this technique with a standard clustering method based on Euclidean distance in $\R^3$. Our method correctly separates the faces of the cube (left) and the legs and the table-top (right). The charts can be useful in many applications such as establishing a $uv$-map for texture mapping and for triangulation based mesh generation. We next describe the latter.

\subsection{Mesh Reconstruction \label{sec:mesh}}
\begin{figure}[H]
    \centering
    \includegraphics[width=\textwidth]{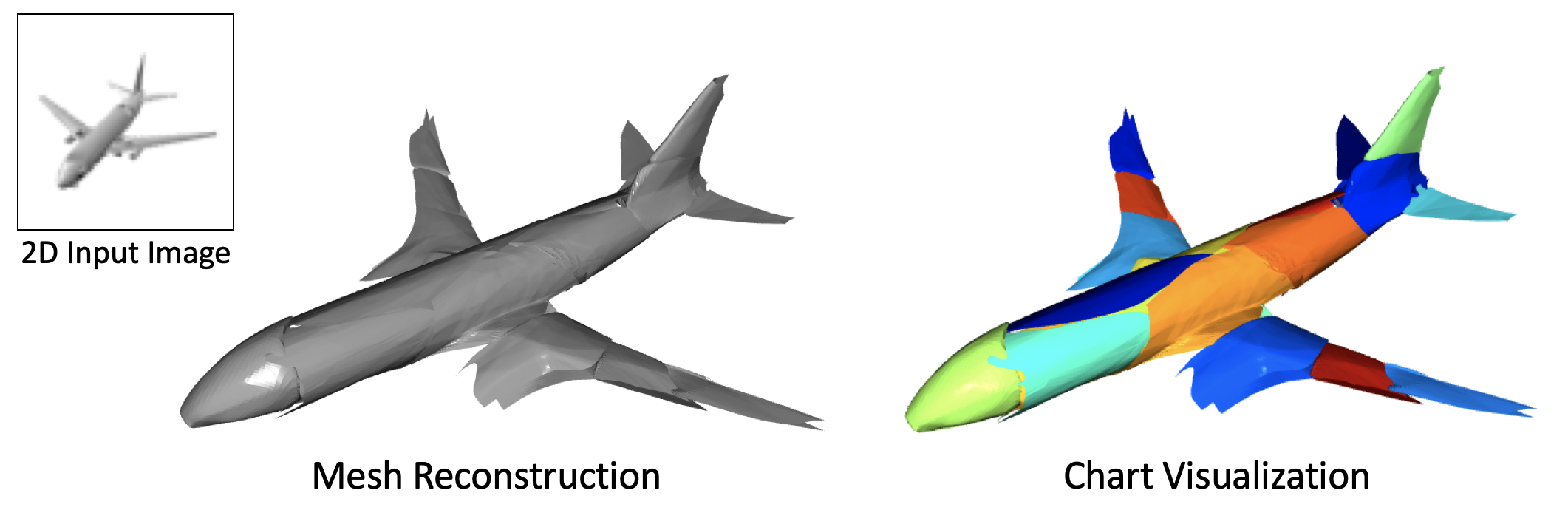}
     \caption{\textbf{Top Left.} 2D Input image. \textbf{Left.} Mesh Reconstruction from \ours{} by fitting a chart onto each non-overlapping manifold from clustering the geodesic lifting coordinates. \textbf{Right.} Visualization of the charts from our decomposition obtained by clustering the geodesic lifting coordinates. Each color indicates a different chart (20 in total). Best viewed in color.}
    \label{fig:mesh_plane}
    \vspace{-10pt}
\end{figure}

Once we decompose the object into charts, we can  fit a surface to each chart and establish a two-dimensional coordinate frame. The 2D coordinates can then be used for triangulation. One possible approach is to fit a parametric representation to each chart such as a plane or conic. In this section, we present a more general approach where we use a multi-layer Perceptron $f_\theta$ to represent the manifold -- similar to the approach presented in AtlasNet. The main advantage of \ours{} over AtlasNet is that since we have low-curvature regions already partitioned into charts and have point coordinates associated with each chart, we can learn the parameters $\theta$ in an unsupervised manner. We present the details of the method in the Supplementary Material, and an illustration of this method in Figure~\ref{fig:mesh_plane}.

\begin{table}[H]
\centering
\begin{tabular}{c | c c c c}
& Pix2Mesh & AtlasNet & OccNet\cite{mescheder2019occupancy} & Ours (Mesh)\\
\hline
Mean Chamfer & 0.216 & 0.175 & 0.215 & \textbf{0.169}\\
Mean Normal & 0.772 & 0.811 & \textbf{0.834} & 0.780\\
\end{tabular}
\vspace{1pt}
\caption{Mesh Reconstruction Comparison on entire ShapeNet test set. Evaluation is done by sampling 100,000 points uniformly on the predicted mesh. The surface normal of each point comes from the surface normal of the triangles the points belong to. Pix2Mesh, AtlasNet and \ours{} are explicit methods whereas OccNet is based on learning occupancy as an implicit function.\label{tab:mesh}}
\vspace{-15pt}
\end{table}
In Figure~\ref{fig:mesh_plane}, we decompose an airplane to 20 non-overlapping charts by clustering the lifting coordinates, triangulating each chart using the method described above. We observe that the charts are split nicely at high-curvature areas. Finally, we compare the resulting meshes with a state of the art implicit function method: Occupancy Network~\cite{mescheder2019occupancy} learns an occupancy function to represent the object and uses marching cubes to generate a mesh. 
In Table~\ref{tab:mesh}, we present mean values over all ShapeNet classes, and present full comparison across classes in the Supplementary Material section. Note that, unlike the other methods in this table, we fit the manifolds as a optimization step rather than directly supervising on the ground truth mesh. Despite the lack of ground truth for generating the mesh, we show competitive Chamfer distance and normal consistency.

\section{Conclusion}

We presented \ours{} for generating continuous surface models from a single image. \ours{} is capable of directly generating surface samples at arbitrary resolution and estimating the geodesic distance between any two samples. Comparisons showed that our method can generate more accurate point samples and surface normals than state of the art methods which can directly generate surfaces. It also performs comparably to implicit function methods which rely on post-processing the output to generate the object surface. We also presented two applications of \ours:~partitioning the surface into non-overlapping, low-curvature charts and learning a a functional mapping $f(u, v)$ of each chart. Combining these two applications allowed us to generate triangulations of the object directly from the network output. This approach alleviates some of the limitations of the previous mesh generation methods such as overlapping charts or genus constraints.

We hope that this work will motivate future research in 3D reconstruction that learns higher-order surface properties like geodesics in order to produce more useful, accurate representations of 3D objects.



\newpage
\section*{Broader Impact}

3D Surface Reconstruction has numerous practical applications in medicine, manufacturing, and entertainment. The ability to reconstruct accurate 3d models from images allows people to transfer knowledge between the physical and virtual worlds. In robotics, 3D reconstruction could assist learning manipulation skills based on the reconstructed objects. In manufacturing and medicine, it can be used for training in a reconstructed virtual environment. While the current machine learning techniques are enabling these beneficial applications, the ability of machines to infer missing information from past experiment remains a double-edge sword. An important question to answer is how machine learning algorithms can identify and handle the biases introduced by the prior experience.

The specific problem of surface reconstruction from a single image is an under-constrained problem: when we are looking at the image of a round object, we can not tell if it is the size of a tennis ball or a planet. Furthermore, we can not be sure that the exact shape is a sphere. Learning based approaches compensate for this lack of knowledge by encoding prior information present in the training dataset. This process might induce biases which can have serious implications (e.g. for organ reconstruction in medical imaging) and reinforce stereotypes (e.g. by inferring a person's 3D model based on their gender or ethnicity) depending on the application. Therefore, practical applications using learning based 3D reconstruction methods must be watchful for such biases and ensure that they are mitigated.
\newpage
{\small
 \bibliographystyle{ieee_fullname}
\bibliography{refs}

\begin{thebibliography}{10}\itemsep=-1pt

\bibitem{balasubramanian2002isomap}
Mukund Balasubramanian and Eric~L Schwartz.
\newblock The isomap algorithm and topological stability.
\newblock {\em Science}, 295(5552):7--7, 2002.

\bibitem{chang2015shapenet}
Angel~X Chang, Thomas Funkhouser, Leonidas Guibas, Pat Hanrahan, Qixing Huang,
  Zimo Li, Silvio Savarese, Manolis Savva, Shuran Song, Hao Su, et~al.
\newblock Shapenet: An information-rich 3d model repository.
\newblock {\em arXiv preprint arXiv:1512.03012}, 2015.

\bibitem{choy2016r2n2}
Christopher~B Choy, Danfei Xu, JunYoung Gwak, Kevin Chen, and Silvio Savarese.
\newblock 3d-r2n2: A unified approach for single and multi-view 3d object
  reconstruction.
\newblock In {\em European Conference on Computer Vision}, pages 628--644.
  Springer, 2016.

\bibitem{choy20163d}
Christopher~B Choy, Danfei Xu, JunYoung Gwak, Kevin Chen, and Silvio Savarese.
\newblock 3d-r2n2: A unified approach for single and multi-view 3d object
  reconstruction.
\newblock In {\em European conference on computer vision}, pages 628--644.
  Springer, 2016.

\bibitem{cox2008multidimensional}
Michael~AA Cox and Trevor~F Cox.
\newblock Multidimensional scaling.
\newblock In {\em Handbook of data visualization}, pages 315--347. Springer,
  2008.

\bibitem{fan2017point}
Haoqiang Fan, Hao Su, and Leonidas~J Guibas.
\newblock A point set generation network for 3d object reconstruction from a
  single image.
\newblock In {\em Proceedings of the IEEE conference on computer vision and
  pattern recognition}, pages 605--613, 2017.

\bibitem{atlasnet}
Thibault Groueix, Matthew Fisher, Vladimir~G Kim, Bryan~C Russell, and Mathieu
  Aubry.
\newblock A papier-m{\^a}ch{\'e} approach to learning 3d surface generation.
\newblock In {\em Proceedings of the IEEE conference on computer vision and
  pattern recognition}, pages 216--224, 2018.

\bibitem{kingma2015adam}
Diederik~P. Kingma and Jimmy Ba.
\newblock Adam: {A} method for stochastic optimization.
\newblock In Yoshua Bengio and Yann LeCun, editors, {\em 3rd International
  Conference on Learning Representations, {ICLR} 2015, San Diego, CA, USA, May
  7-9, 2015, Conference Track Proceedings}, 2015.

\bibitem{lin2018learning}
Chen-Hsuan Lin, Chen Kong, and Simon Lucey.
\newblock Learning efficient point cloud generation for dense 3d object
  reconstruction.
\newblock In {\em AAAI Conference on Artificial Intelligence ({AAAI})}, 2018.

\bibitem{Luciano2018}
Lorenzo Luciano and A.~Ben Hamza.
\newblock Deep learning with geodesic moments for 3d shape classification.
\newblock {\em Pattern Recognition Letters}, 105:182--190, Apr. 2018.

\bibitem{maaten2008visualizing}
Laurens van~der Maaten and Geoffrey Hinton.
\newblock Visualizing data using t-sne.
\newblock {\em Journal of machine learning research}, 9(Nov):2579--2605, 2008.

\bibitem{mescheder2019occupancy}
Lars Mescheder, Michael Oechsle, Michael Niemeyer, Sebastian Nowozin, and
  Andreas Geiger.
\newblock Occupancy networks: Learning 3d reconstruction in function space.
\newblock In {\em Proceedings of the IEEE Conference on Computer Vision and
  Pattern Recognition}, pages 4460--4470, 2019.

\bibitem{mitchell2019higher}
Eric Mitchell, Selim Engin, Volkan Isler, and Daniel~D Lee.
\newblock Higher-order function networks for learning composable 3d object
  representations.
\newblock {\em arXiv preprint arXiv:1907.10388}, 2019.

\bibitem{pai2019dimal}
G. {Pai}, R. {Talmon}, A. {Bronstein}, and R. {Kimmel}.
\newblock Dimal: Deep isometric manifold learning using sparse geodesic
  sampling.
\newblock In {\em 2019 IEEE Winter Conference on Applications of Computer
  Vision (WACV)}, pages 819--828, 2019.

\bibitem{park2019deepsdf}
Jeong~Joon Park, Peter Florence, Julian Straub, Richard Newcombe, and Steven
  Lovegrove.
\newblock Deepsdf: Learning continuous signed distance functions for shape
  representation.
\newblock {\em arXiv preprint arXiv:1901.05103}, 2019.

\bibitem{riegler2017octnet}
Gernot Riegler, Ali Osman~Ulusoy, and Andreas Geiger.
\newblock Octnet: Learning deep 3d representations at high resolutions.
\newblock In {\em The IEEE Conference on Computer Vision and Pattern
  Recognition (CVPR)}, July 2017.

\bibitem{tatarchenko2017octree}
Maxim Tatarchenko, Alexey Dosovitskiy, and Thomas Brox.
\newblock Octree generating networks: Efficient convolutional architectures for
  high-resolution 3d outputs.
\newblock In {\em Proceedings of the IEEE International Conference on Computer
  Vision}, pages 2088--2096, 2017.

\bibitem{tenenbaum2000global}
Joshua~B Tenenbaum, Vin De~Silva, and John~C Langford.
\newblock A global geometric framework for nonlinear dimensionality reduction.
\newblock {\em science}, 290(5500):2319--2323, 2000.

\bibitem{wang2018pixel2mesh}
Nanyang Wang, Yinda Zhang, Zhuwen Li, Yanwei Fu, Wei Liu, and Yu-Gang Jiang.
\newblock Pixel2mesh: Generating 3d mesh models from single rgb images.
\newblock In {\em Proceedings of the European Conference on Computer Vision
  (ECCV)}, pages 52--67, 2018.

\bibitem{wang2000shape}
Y. {Wang}, B.~S. {Peterson}, and L.~H. {Staib}.
\newblock Shape-based 3d surface correspondence using geodesics and local
  geometry.
\newblock In {\em Proceedings IEEE Conference on Computer Vision and Pattern
  Recognition. CVPR 2000 (Cat. No.PR00662)}, volume~2, pages 644--651 vol.2,
  2000.

\bibitem{Wang2003}
Yongmei Wang, Bradley~S. Peterson, and Lawrence~H. Staib.
\newblock 3d brain surface matching based on geodesics and local geometry.
\newblock {\em Computer Vision and Image Understanding}, 89(2-3):252--271, Feb.
  2003.

\bibitem{wu2016learning}
Jiajun Wu, Chengkai Zhang, Tianfan Xue, Bill Freeman, and Josh Tenenbaum.
\newblock Learning a probabilistic latent space of object shapes via 3d
  generative-adversarial modeling.
\newblock In D.~D. Lee, M. Sugiyama, U.~V. Luxburg, I. Guyon, and R. Garnett,
  editors, {\em Advances in Neural Information Processing Systems 29}, pages
  82--90. Curran Associates, Inc., 2016.

\bibitem{yang2018foldingnet}
Yaoqing Yang, Chen Feng, Yiru Shen, and Dong Tian.
\newblock Foldingnet: Point cloud auto-encoder via deep grid deformation.
\newblock In {\em Proceedings of the IEEE Conference on Computer Vision and
  Pattern Recognition}, pages 206--215, 2018.

\end{thebibliography}
}
\newpage
\appendix
\setcounter{section}{0}

{\Large{\bf{
Supplementary Material}
}}


\section{\ours{} Training Details}

\subsection{Network Architecture}
\begin{figure} [H]
    \centering
    \includegraphics[width=\textwidth]{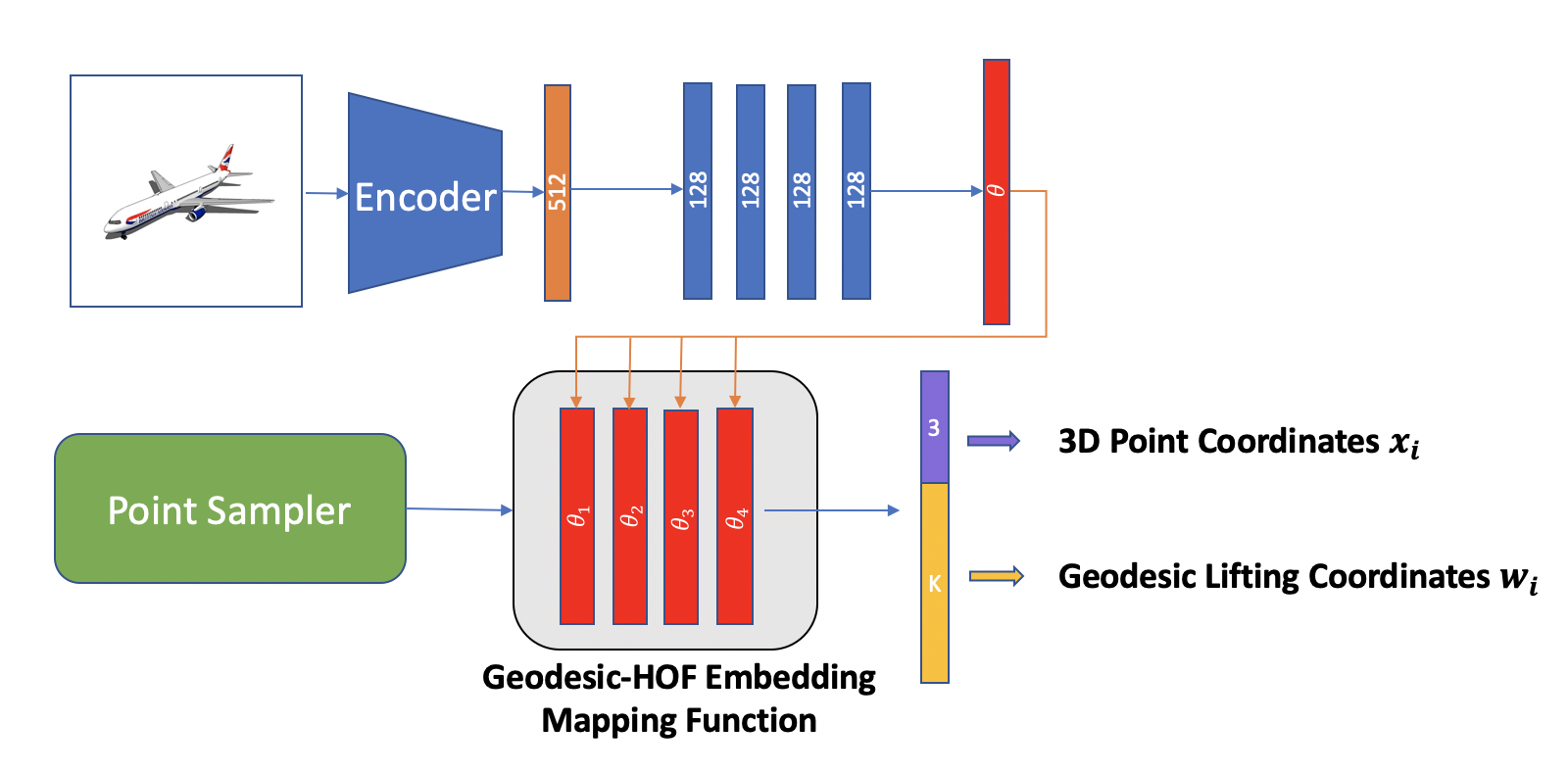}
    \caption{Architecture overview of \ours{}}
    \label{fig:arch}
\end{figure}

Figure~\ref{fig:arch} illustrates the main architecture of \ours{}. An image encoder takes a single-view image of the 3D object to be constructed and maps it to a 512-dimensional image feature vector. The Higher-Order Function network uses this feature descriptor as input and outputs $\theta$ -- the weights of the mapping function $f$. 
The mapping network $f$ maps each samples from $\R^3$ to an point in  $3+K$-dimensional embedding space. The first three dimensions of the embedding vector, as denoted with purple in the figure are the ''3D Point Coordinates $x_i$", which represent the true 3D location of these samples in the object space. The next $K$ dimensions are called "Geodesic Lifting Coordinates $w_i$". 


\subsection{Dataset and Training}

In order to give a fair comparison of \ours{} with other methods, we use the same evaluation procedure as stated in Occupancy Networks ~\cite{mescheder2019occupancy}. We use the pre-processed ShapeNet~\cite{chang2015shapenet} where the objects are pre-processed so that the object surface is  water-tight. We use the same split with 30661 training objects, 4371 validation objects and 8751 test objects. The objects are normalized so that the minimum bounding box of the shape is a unit square. The Chamfer numbers are reported with $1/10$ of each object bounding box as the unit for chamfer. For the details of the this data-processing, please refer to~\cite{mescheder2019occupancy}. We train our network on a single Gtx-1080Ti GPU with an Adam Optimizer with learning rate 1e$-5$ as the learning rate. We train the network for 5 epochs, where in each epoch we go through all 24 images of all objects. Batch size for training is 1.

\newpage
\section{Explicit Function Representation of Charts}

In order to reconstruct the object surface from our predicted embedding $Z = \{z_i \in \R^{3+K}\}$, we simply optimize a set of mappings from a canonical chart to different manifolds that each consistent of a sub-region of the object's surface. For each decomposed manifold, we want to learn a mapping from a canonical chart to this manifold. More specifically, given a set of samples $U_i = \{(u, v) \in \R^2\}$ from a canonical 2d chart and a set of samples from target manifold $M_i$, we want to learn the mapping function $f: \R^2 \rightarrow \R^3$ so as to minimize the Chamfer distance between $U_i$ and $M_i$. In practice, we decompose the object surface into 20 local manifolds (charts) such that $M = \bigcup\limits_{i=1}^{20} M_{i}$. 

This manifold fitting methodology is similar to AtlasNet~\cite{atlasnet} with a number of critical differences:
\begin{enumerate}
    \item We are not optimizing the mesh with respect to the ground truth during training. The optimization is with respect to our own prediction given by the 3D point coordinates predicted by our network.
    \item There is zero to little  overlap between the charts since the points are partitioned using geodesic clustering before fitting.
    \item The decomposition with the geodesic lifting dimensions simplifies the manifold learning problem into learning the mapping to a set of low-curvature manifolds.
\end{enumerate}
Rather than designing a new mesh learning method, we show that a surface can be easily extracted from our embedding representation and that we can achieve competitive mesh results. In practice, one can imagine extending this to a fully end-to-end mesh reconstruction algorithm to directly supervise the network with the ground truth training set. Yet this is not the focus of this paper and we leave the exploration of this feature in our future work.

For fitting the mesh, we represent each mapping function $f_i$ as a neural network that takes in samples from an unit square chart and maps it to a low-curvature area of the object. We optimize the the Chamfer loss between the mapped samples and the ground truth object samples. In practice, we take 100 gradient steps for all of the charts of each object to obtain a nice-looking mesh. Note that this is done purely as an optimization step on the output of the network, we are mapping a set of charts onto our predicted point set. We notice the quality of the mesh grows as we take more gradient steps.

\section{Qualitative Mesh Evaluation}
\begin{figure}
    \centering
    \includegraphics[width=0.95\linewidth]{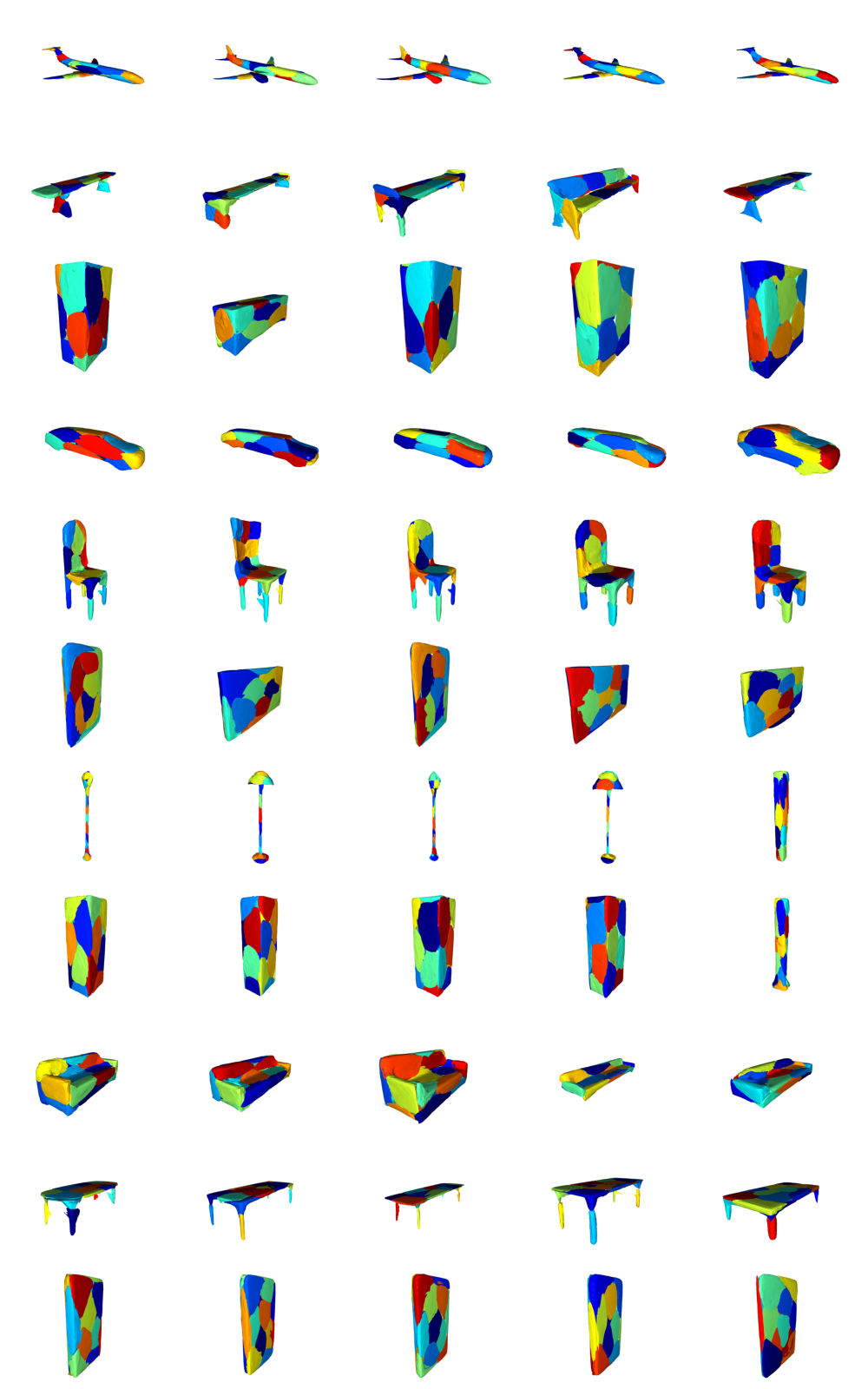}
    \caption{Mesh rendering of select classes. Each row shows the objects from a class in ShapeNet~\cite{chang2015shapenet} Test set. Color indicates the label of each manifold. We use 20 non-overlapping manifolds to represent the object.}
    \label{fig:all_supp_color}
\end{figure}

We  present the qualitative rendering of the meshes from each object category in Figure~\ref{fig:all_supp_color}. Similar to what we present in Figure~\ref{fig:mesh_plane}, each color here indicates a different component from our manifold decomposition. As seen in  Figure~\ref{fig:all_supp_color}, each sub-manifold generally corresponds to a low-curvature manifold that can be learned more easily. The geodesic lifting dimensions of the \ours{} are used to identify these regions. Due to the clustering nature of our decomposition, in which a point is assigned to only one cluster ID, we have little to no overlap between any two manifolds. Since the focus of the paper is not mesh reconstruction, we did not explicitly use the decomposition in learning the mapping function. However, we believe the near-mutual exclusiveness between the manifolds could be a desirable property in mesh reconstruction. Specifically, this property allows the network to avoid learning repeated mapping between charts, which enables more efficient use of the network capacity.
\newpage

\section{Full Mesh Comparison}
In this section, we show the full quantitative evaluation of \ours{} compared with the mesh-based methods reported in the main paper. As shown in Table~\ref{tab:full_chamfer}, \ours{} is able to learn very accurate point prediction while learning the geodesic embedding. In terms of Chamfer performance, \ours{} achieves the best results among all mesh-based methods in this table. In Normal Consistency, we can see a clear advantage of implicit methods over direct mapping methods. We hypothesize that this is because the marching cubes post-processing inherently smooths the object surface and produces more robust normal prediction. We observe a trade-off between detailed features (Chamfer) and smoothness of surface prediction (normal). Note that, unlike the other methods in this table, our mesh algorithm is a simple post-processing of the output of \ours{}, and therefore does not have access to the ground truth training set. 
\begin{table}
    \centering
    \setlength{\tabcolsep}{3pt}
    \begin{tabular}{|c | c c c c|}
    \hline
    Category  & Pix2Mesh~\cite{wang2018pixel2mesh} & AtlasNet~\cite{atlasnet} & OccNet~\cite{mescheder2019occupancy} & Ours (Mesh)\\
    \hline
    Airplane    & 0.187 &               \textbf{0.104}          & 0.147 &            0.112\\
    Bench       & 0.201 &               0.138                   & 0.155 &            \textbf{0.122}\\
    Cabinet     & 0.196 &               0.175                   & 0.167 &            \textbf{0.158}\\
    Car         & 0.180 &               0.141                   & 0.159 &            \textbf{0.120}\\
    Chair       & 0.265 &               0.209                   & 0.228 &            \textbf{0.204}\\
    Display     & 0.239 &               \textbf{0.198}          & 0.278 &            0.213\\
    Lamp        & 0.308 &               0.305                   & 0.479 &            \textbf{0.280}\\
    Speaker     & 0.285 &               0.245                   & 0.300 &            \textbf{0.239}\\
    Rifle       & 0.164 &               \textbf{0.115}          & 0.141 &            0.117\\
    Sofa        & 0.212 &               0.177                   & 0.194 &            \textbf{0.176}\\
    Table       & 0.218 &               0.190                   & \textbf{0.189}&    0.193\\
    Telephone   & 0.149 &               0.128                   & 0.140 &            \textbf{0.115}\\
    Vessel      & 0.212 &               \textbf{0.151}          & 0.218 &            0.157\\
    \hline
    mean &  0.216 & 0.175 & 0.215 & \textbf{0.169}\\
    \hline
    \end{tabular}
    \vspace{5pt}
    \caption{Chamfer Comparison: \ours{} achieves state-of-the art performance in Chamfer distance. Overall and in certain categories, \ours{} significantly outperforms the competing methods. We sample 100,000 points on the object of interest and the output of each method to compute the Chamfer distance.}
    \label{tab:full_chamfer}
\end{table}

\begin{table}
    \centering
    \setlength{\tabcolsep}{3pt}
    \begin{tabular}{|c|c c c c|}
    \hline
    Category & Pix2Mesh~\cite{wang2018pixel2mesh} & AtlasNet~\cite{atlasnet} & OccNet~\cite{mescheder2019occupancy} & Ours (Mesh) \\
    \hline
    Airplane    & 0.759 & 0.836              & \textbf{0.840}               & 0.805            \\
    Bench       & 0.732 & 0.779              & \textbf{0.813}      & 0.766            \\
    Cabinet     & 0.834 & 0.850              & \textbf{0.879}      & 0.833            \\
    Car         & 0.756 & 0.836              & \textbf{0.852}      & 0.775            \\
    Chair       & 0.746 & 0.791              & \textbf{0.823}      & 0.762            \\
    Display     & 0.830 & \textbf{0.858}     & 0.854               & 0.840            \\
    Lamp        & 0.666 & 0.694              & \textbf{0.731}               & 0.657            \\
    Speaker     & 0.782 & 0.825              & \textbf{0.832}               & 0.780            \\
    Rifle       & 0.718 & 0.725              & \textbf{0.766}               & 0.727            \\
    Sofa        & 0.820 & 0.840              & \textbf{0.863}      & 0.826            \\
    Table       & 0.784 & 0.832              & \textbf{0.858}      & 0.800            \\
    Telephone   & 0.907 & 0.923              & \textbf{0.935}      & 0.910            \\
    Vessel      & 0.699 & 0.756              & \textbf{0.794}      & 0.725            \\
    \hline
    mean &  0.772 & 0.811 & \textbf{0.834} & 0.780\\
    \hline
    \end{tabular}
    \vspace{5pt}
    \caption{Normal Comparison: \ours{} performs competitively against direct mapping methods such as Pix2Mesh~\cite{wang2018pixel2mesh} and AtlasNet~\cite{atlasnet}. An implicit surface method, Occupancy Networks~\cite{mescheder2019occupancy}, outperforms all direct mapping methods in this table. Note that \ours{} post-processes the network output and does not have access to the ground truth in the training set.}
\end{table}
\end{document}